\documentclass[conference]{IEEEtran}
\IEEEoverridecommandlockouts
\usepackage{cite}
\usepackage{amsmath,amssymb,amsfonts}
\usepackage{algorithm, algorithmic}
\usepackage{graphics, graphicx, pgfplots}
\usepackage{textcomp, booktabs}
\usepackage{xcolor}
\def\BibTeX{{\rm B\kern-.05em{\sc i\kern-.025em b}\kern-.08em
    T\kern-.1667em\lower.7ex\hbox{E}\kern-.125emX}}
\begin{document}

\title{Condensed Composite Memory Continual Learning}

\author{\IEEEauthorblockN{Felix Wiewel and Bin Yang}
	\IEEEauthorblockA{\textit{Institute of Signal Processing and System Theory} \\
		\textit{University of Stuttgart}\\
		Stuttgart, Germany \\
		\{felix.wiewel, bin.yang\}@iss.uni-stuttgart.de}
}

\maketitle

\begin{abstract}

Deep Neural Networks (DNNs) suffer from a rapid decrease in performance when trained on a sequence of tasks where only data of the most recent task is available. This phenomenon, known as catastrophic forgetting, prevents DNNs from accumulating knowledge over time. Overcoming catastrophic forgetting and enabling continual learning is of great interest since it would enable the application of DNNs in settings where unrestricted access to all the training data at any time is not always possible, e.g. due to storage limitations or legal issues. While many recently proposed methods for continual learning use some training examples for rehearsal, their performance strongly depends on the number of stored examples. In order to improve performance of rehearsal for continual learning, especially for a small number of stored examples, we propose a novel way of learning a small set of synthetic examples which capture the essence of a complete dataset. Instead of directly learning these synthetic examples, we learn a weighted combination of shared components for each example that enables a significant increase in memory efficiency. We demonstrate the performance of our method on commonly used datasets and compare it to recently proposed related methods and baselines.

\end{abstract}


\section{Introduction}

DNNs have achieved remarkable results on many challenging problems, e.g. computer vision, but still struggle on some important aspects when compared with the way humans learn. One crucial aspect where DNNs fall short is the continuous accumulation of knowledge when trained on a sequence of disjoint tasks. Exposed to such a setting where only the training data of the most recent task is available, DNNs suffer from the phenomenon of catastrophic forgetting \cite{mccloskey1989catastrophic, french1999catastrophic, lewandowsky1995catastrophic}, i.e. a rapid decrease in performance on previously learned tasks whenever a new task is trained. Although this phenomenon has been known since the early $1990$s, it has only recently seen renewed interest in the scientific community in the form of many publications on methods to overcome catastrophic forgetting and some studies of the phenomenon itself. Solving the challenges associated with it would benefit a lot of applications and DNNs in general. One could move away from the current paradigm where all data must be accessible before training of a DNN can begin on to a setting where it can accumulate knowledge over time. Not only could this enable autonomous agents that continually learn from their environment but also applications where legal or privacy aspects prohibit the accumulation of large data sets of personal data. In the latter case, a spatial aggregation of all private data in one place could be replaced with a temporal aggregation through a DNN that is sequentially trained on small isolated parts of the whole dataset. Applications like these and the general advancement of DNNs and their abilities make the study of catastrophic forgetting and how to overcome it, also known as continual, continuous or life long learning in the literature, a relevant and interesting field in the area of artificial intelligence. Many different methods for continual learning published in the recent years can be roughly classified in three distinct categories.

\textbf{I)} Regularization based methods use some sort of loss term for a DNN in order to consolidate the knowledge acquired on previously learned classes. Canonical examples for this class of methods are \textit{Elastic Weight Consolidation} (EWC)\cite{kirkpatrick2017overcoming} and \textit{Synaptic Intelligence} (SI)\cite{zenke2017continual}. Both use a weighted quadratic loss in order to penalize deviations from previously learned parameters. While EWC uses an estimate of the Fisher Information, SI uses the movement of a DNNs parameters and corresponding decrease in loss during training in order to measure the importance of a parameter to previously learned tasks.

\textbf{II)} Structural methods change the structure of a DNN in order to enable continual learning either by expanding it or by selectively disabling parts of the model in order to avoid interference between neural activities of different tasks. \textit{Progressive Neural Networks} (PNN)\cite{rusu2016progressive} expand a DNN by adding new layers and lateral connections between those and frozen ones that were trained on previous tasks. As the number of lateral connections increases dramatically when many new layers are added, scaling this method to bigger models is challenging. The only recently proposed method \textit{A Neuromodulated Meta-Learning Algorithmn} (ANML)\cite{beaulieu2020learning} instead uses meta learning and a neuromodulatory network in order to gate inputs of a layer conditioned on the input.

\textbf{III)} Rehearsal with a small set of training examples is another common strategy for avoiding catastrophic forgetting. There exist several methods that use this approach in the literature. They mainly differ in the type of examples used for rehearsal, i.e. generated \cite{shin2017continual, DBLP:journals/corr/abs-1809-10635}, synthetic \cite{zhao2020dataset, wang2018dataset} or stored images, in the way examples are selected \cite{borsos2020coresets} and in the position at which the rehearsal data is injected into the model\cite{pellegrini2019latent} during training. Most methods, with the exception of generative replay, have the nice property that their performance scales with the number of examples that are stored for rehearsal. While \textit{Bias Correction} (BiC)\cite{wu2019large} and \textit{Remind}\cite{hayes2020remind} have shown that rehearsal can be scaled to a large scale, they store a significant amount of the training examples of the original dataset. Especially \textit{Remind} stores roughly $960$ thousand of the $1.2$ million images of the Imagenet dataset in a compressed form. Although this is a viable way of increasing the performance of a rehearsal based method, it raises the question if there are more elegant ways than just trying to store more examples by means of an ever increasing compression ratio.

In this paper, we follow an alternative paradigm by learning synthetic examples for rehearsal directly in a compressed form. We base our approach\footnote{Code available at https://github.com/FelixWiewel/CCMCL} on the inspiring work of Zhao et. al. \cite{zhao2020dataset} which uses a novel gradient matching approach for condensation of a large dataset into a small number of synthetic examples. Instead of naively using dataset condensation to replace randomly selected samples in rehearsal, we propose an extension to the original method that is specifically designed for continual learning. We propose a novel way of how condensed examples are represented in order to further increase the memory efficiency. For this we use a weighted combination of learned components to represent each example. This enables us to share common features across different examples and therefore increase the memory efficiency significantly when compared to the original dataset condensation algorithm. Finally we demonstrate our methods performance on commonly used benchmarks for continual learning and compare it to recently proposed related methods and baselines.

\section{Incremental class learning}
\label{sec:IncrementalClassLearning}

While there are three commonly considered scenarios in the literature on continual learning, i.e. incremental task, incremental domain and incremental class learning, we focus on the latter one as it is the most challenging one. The reason for this is the lack of a task identifier that is commonly exploited in the other, less challenging, scenarios. In those the task identifier is commonly used to select a separate output layer for each task or domain. Since this layer then only has to solve the selected task, it effectively reduces the problem to a simpler one, e.g. a binary classification. While there might be a task identifier available at test time in some applications, it is not available in general and therefore restricts any method exploiting it to a limited number of applications. In incremental class learning, on the other hand, there is no task identifier and the DNN has only one output layer that is expanded with each newly learned class. In contrast to the scenarios where a task identifier is available, the DNN therefore has to solve a problem that becomes increasingly difficult with every added class. For detailed definitions of all scenarios and their challenges we refer the reader to \cite{hsu2018re, van2019three}.

In incremental class learning, a sequence of $N_{T}$ classification tasks $\mathcal{T}_{1\leq i\leq N_{T}}=\lbrace\mathbf{x}_{j}^{i},y_{j}^{i}\vert1\leq j\leq M_{i}\rbrace$, each represented by a set of $M_{i}$ input-output pairs, is considered. The $j$-th input of the $i$-th task $\mathbf{x}_{j}^{i}\in\mathbb{R}^{S}$ is given by a real-valued vector, e.g. the $S$ pixels of an image, and its corresponding target value $y_{j}^{i}\in\mathbb{Z}$ is an integer class label. These tasks typically contain only examples of two or more classes that are not present in any other task. A standard technique to construct these is to split an image classification dataset into disjoint sets. Commonly used examples are the MNIST \cite{lecun2010mnist}, FashionMNIST \cite{DBLP:journals/corr/abs-1708-07747}, SVHN \cite{Netzer2011} and CIFAR10 \cite{Krizhevsky09learningmultiple} datasets that are split into five task containing two classes in increasing order, i.e. $y_{j}^{1}\in\lbrace0,1\rbrace$, $y_{j}^{2}\in\lbrace2,3\rbrace$, $\ldots$, $y_{j}^{5}\in\lbrace8,9\rbrace$. The resulting datasets are called SplitMNIST, SplitFashionMNIST, SplitSVHN and SplitCIFAR10. Given these task definitions, the goal in incremental class learning is to train a DNN $f_{\boldsymbol{\theta}}(\mathbf{x})$ on the sequence $\mathcal{T}_{1},\mathcal{T}_{2},\ldots,\mathcal{T}_{N_{T}}$ such that an empirical risk\cite{vapnik1992principles} $\mathcal{L}$ is minimized over all of them once training on the sequence is complete. Formally this can be written as
\begin{align}
\boldsymbol{\theta}^{\star}=\arg\min_{\boldsymbol{\theta}}\sum_{i=1}^{N_{T}}\sum_{j=1}^{M_{i}}\mathcal{L}(f_{\boldsymbol{\theta}}(\mathbf{x}_{j}^{i}),y_{j}^{i}),
\end{align}
where $\mathcal{L}:\mathbb{R}^{S}\times\mathbb{Z}\rightarrow\mathbb{R}$ is typically chosen as the cross-entropy loss between the DNNs prediction $f_{\boldsymbol{\theta}}(\mathbf{x}_{j}^{i})$ and a one-hot encoding of the class label $y_{j}^{i}$. This setup is different from a standard learning problem as during the training on task $\mathcal{T}_{i}$ only its corresponding dataset is available to the DNN. This leads to two major challenges that have to be overcome for continual learning. First, the DNN has to be stable enough during training such that the knowledge learned on previous tasks $\mathcal{T}_{1},\ldots,\mathcal{T}_{i-1}$ is retained. Additionally, it has to be plastic enough to learn additional concepts of the newly trained tasks. Satisfying both of these contradicting requirements at the same time is known as the stability-plasticity dilemma in the literature\cite{Carpenter:87}. Since DNNs tend to be very plastic, i.e. they can approximate almost any function with a large enough number of parameters rather quickly, they suffer from catastrophic forgetting and typically need additional measures to ensure stability.

\section{Proposed Method}

As our method is based on dataset condensation, we first introduce the basic idea and gradient matching as a suitable loss in section \ref{ssec:DatasetCondensation}. We then proceed to present the novel approach of a composite memory in section \ref{ssec:CompositeMemory} and finally summarize our method in \ref{ssec:Algorithm}.

\subsection{Dataset Condensation}
\label{ssec:DatasetCondensation}

Consider a given dataset $\mathcal{D}=\lbrace\mathbf{x}_{i}^{D},y_{i}^{D}\rbrace_{i=1}^{M}$ of $M$ inputs $\mathbf{x}_{i}\in\mathbb{R}^{S}$ and their corresponding targets $\mathbf{y}_{i}\in\mathbb{Z}$, a DNN $f_{\boldsymbol{\theta}}(\mathbf{x})$ with parameters $\boldsymbol{\theta}$ and an empirical risk $\mathcal{L}(f_{\boldsymbol{\theta}}(\mathbf{x}),y)\in\mathbb{R}$. The goal of dataset condensation is to find a small set of synthetic input-output pairs $\mathcal{S}=\lbrace\mathbf{x}_{i}^{S},y_{i}^{S}\rbrace_{i=1}^{N}$, where $N\ll M$, such that the optimal parameters
\begin{align}
\boldsymbol{\theta}_{D}^{\star}=\arg\min_{\boldsymbol{\theta}}\sum_{i=1}^{M}\mathcal{L}(f_{\boldsymbol{\theta}}(\mathbf{x}_{i}^{D}),y_{i}^{D})
\end{align}
found on the original dataset are similar to the optimal parameters 
\begin{align}
\boldsymbol{\theta}_{S}^{\star}=\arg\min_{\boldsymbol{\theta}}\sum_{i=1}^{N}\mathcal{L}(f_{\boldsymbol{\theta}}(\mathbf{x}_{i}^{S}),y_{i}^{S})
\end{align}
obtained using the synthetic dataset $\mathcal{S}$. Depending on how this similarity is measured, different algorithms for obtaining the small synthetic dataset $\mathcal{S}$ can be derived. The authors of \cite{wang2018dataset} propose to learn the synthetic examples and a learning rate in such a way that one gradient decent step on them leads to a high performance on the original test set. In order to find a solution that generalizes well on many different random initializations of the DNN parameters $\boldsymbol{\theta}$, they further propose to minimize the expected empirical risk when training on $\mathcal{S}$ where the expectation is taken over many different random initializations. In their inspiring work \cite{zhao2020dataset}, Zhao et. al. propose two different ways of measuring the aforementioned similarity. First they argue that minimizing the distance in parameter space is a possible but computationally expensive method of learning $\mathcal{S}$ as it results in a bi-level optimization problem where the inner loop involves training a potentially large-scale model many times. In order to avoid this they propose to exploit the dynamics during training on both the original dataset $\mathcal{D}$ and the synthetic one $\mathcal{S}$ by matching their gradients. The intuition behind this is that the trajectory of $\boldsymbol{\theta}$ through the parameter space when trained on $\mathcal{S}$ should be similar to a training on the original dataset $\mathcal{D}$. This can be enforced by
\begin{align}
\label{eq:grad_matching}
\min_{\mathcal{S}}\mathbb{E}_{\boldsymbol{\theta}_{0}\sim P_{\boldsymbol{\theta}_{0}}}\left[\sum_{t=1}^{T}D(\nabla_{\boldsymbol{\theta}}\mathcal{L}_{\mathcal{D}}(\boldsymbol{\theta}_{t}),\nabla_{\boldsymbol{\theta}}\mathcal{L}_{\mathcal{S}}(\boldsymbol{\theta}_{t}))\right],
\end{align}
where the expectation is taken over many different random initializations of $\boldsymbol{\theta}$ following the distribution $P_{\boldsymbol{\theta}_{0}}$, $D(\cdot,\cdot)$ is some distance measure, and $\nabla_{\boldsymbol{\theta}}\mathcal{L}_{\mathcal{D}}(\boldsymbol{\theta}_{t})$ as well as $\nabla_{\boldsymbol{\theta}}\mathcal{L}_{\mathcal{S}}(\boldsymbol{\theta}_{t}))$ are gradients of the empirical risk w.r.t. $\boldsymbol{\theta}$ on the original and synthetic datasets $\mathcal{D}$ and $\mathcal{S}$ at training step $t$. For a more detailed description on how to arrive at Eq. \ref{eq:grad_matching} the reader is referred to \cite{zhao2020dataset}. While there are many possible distance measures $D(\cdot,\cdot)$, Zhao et. al. propose
\begin{align}
D(\mathbf{A},\mathbf{B})=\sum_{i=1}^{out}\left(1-\dfrac{\mathbf{A}_{i}\cdot\mathbf{B}_{i}}{\Vert\mathbf{A}_{i}\Vert\Vert\mathbf{B}_{i}\Vert}\right),
\end{align}
where $\mathbf{A}_{i}$ and $\mathbf{B}_{i}$ are flattened vectors of the weights corresponding to the $i$-th output node and $\Vert\cdot\Vert$ is the euclidean norm. For a fully connected layer with a weight matrix $\mathbf{W}\in\mathbb{R}^{out\times in}$, this results in the cosine similarity between the gradients of individual output neurons. For a convolutional layer with a kernel tensor $\mathbf{W}\in\mathbb{R}^{h\times w\times in\times out}$, a simple assignment to individual neurons is not possible. It is also interesting to note that the contribution of biases for both fully connected and convolutional layers is ignored. But similarly to Zhao et. al. we found that this distance enables the usage of a single learning rate for all layers in a network and therefore we adopt it for all our experiments.

\subsection{Composite Memory}
\label{ssec:CompositeMemory}

\begin{figure}
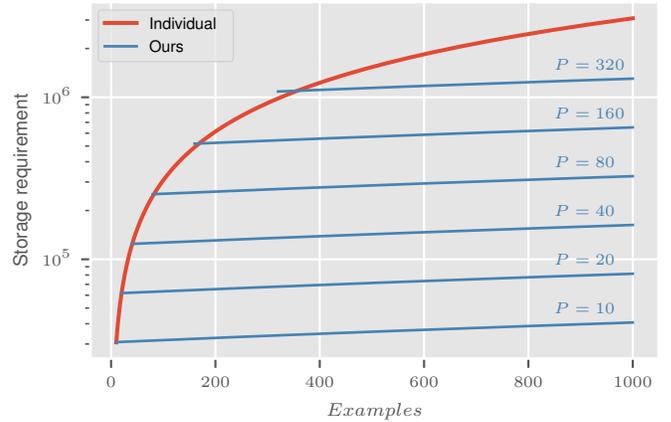

	\include{Figures/plot_memory}
	\vspace{-0.75cm}
	\caption{Storage requirement for $32\times 32$ RGB-Images of the proposed composite memory for different values of $P$ and individual storing. Note the logarithmic scale of the Y-axis.}
	\label{fig:memory}
\end{figure}

Although dataset condensation using gradient matching as shown in Eq. \ref{eq:grad_matching} yields quite impressive results, its memory efficiency can be improved by exploiting similarities between examples of one particular class. For this, we try to decompose the learned synthetic data into a fixed number of shared components. We base this approach on the observation that instances of one specific class typically posses common features that can be shared among them and therefore only need to be stored once. Individual examples can then be reconstructed from a  weighted combination of these learned common features. Formally we propose to represent each example $\mathbf{x}_{j}$ as a function
\begin{align}
\label{eq:composite}
\mathbf{x}_{j}=\sigma\left(\sum_{i=1}^{P}w_{ij}\mathbf{c}_{i}\right),
\end{align}
where $w_{ij}\in\mathbb{R}$ is the mixing coefficient of the $i$-th component $\mathbf{c}_{i}\in\mathbb{R}^{S}$ and $\sigma:\mathbb{R}^{S}\rightarrow\mathbb{R}^{S}$ is the element-wise applied Sigmoid function. This allows us to learn these shared features and their contributions to every example separately. In contrast to the original dataset condensation proposed in \cite{zhao2020dataset}, we can therefore significantly increase the memory efficiency by sharing features among the stored examples instead of explicitly saving them multiple times. In order to quantify this, we consider an example of storing $Q$ $S$-dimensional examples independently as proposed in \cite{zhao2020dataset} and according to Eq. \ref{eq:composite}. Storing them individually requires $B_{ind}=Q\times S$ real numbers. Assuming $P$ shared features $\mathbf{c}_{i}$ our method requires storing $B_{cm}=P\times(S+Q)$ real numbers. If $S\gg Q$, which typically holds true for high dimensional inputs and a small number of examples stored for rehearsal, the storage requirement $B_{cm}$ of our method is approximately constant while $B_{ind}$ grows linearly with the number of stored examples $Q$. Fig. \ref{fig:memory} visualizes the storage requirement of these two approaches in case of the CIFAR10 dataset. Even though it contains relatively small RGB images with $32\times 32$ pixels the storage requirement of our method remains approximately constant as the number of stored examples $Q$ increases.

While our method of representing examples by a weighted combination of components is similar to the well known Principal Component Analysis (PCA)\cite{pearson1901liii}, it is fundamentally different in the way components are learned. Rather than trying to find a representation of the data that captures most of the variance in it, we use the gradient matching loss from Eq. \ref{eq:grad_matching} in order to learn the components to match the gradients of an original training dataset. This effectively combines the data compression aspect of the PCA with dataset condensation and enables learning a set of synthetic training examples in a compressed form. Although this can improve performance, it also adds new hyper parameters, the number of learned components $\mathbf{c}_{i}$ and weights $w_{ij}$, to the dataset condensation algorithm which need to be selected. How we select them in this work is described in section \ref{ssec:Hyperparameters}.

\begin{algorithm}
	\begin{algorithmic}[1]
		\REQUIRE Buffer $\mathcal{B}$, task $\mathcal{T}_{i}$, composite memory $\mathcal{M}$, training optimizer $opt_{T}$, condensation optimizer $opt_{C}$, training batch size $B_{T}$, condensation batch size $B_{C}$, training learning rate $\gamma_{T}$, condensation learning rate $\gamma_{C}$, outer iterations $K$, inner iterations $T$, matching iterations $I$, training iterations $J$, DNN $f_{\boldsymbol{\theta}}$
		\ENSURE Updated buffer $\mathcal{B}$
		\STATE Initialize composite memory $\mathbf{c}_{i}$ and $w_{ij}$
		\FOR{$K$ iterations}
		\STATE Reinitialize model $f_{\boldsymbol{\theta}}$
		\FOR{$T$ iterations}
		\STATE Sample from one class of task $\mathcal{B}_{T}\sim\mathcal{T}_{i}$
		\STATE Sample from composite memory $\mathcal{B}_{M}\sim\mathcal{M}$
		\STATE Compute gradient $\mathbf{g}_{T}=\nabla_{\boldsymbol{\theta}}\sum_{\mathbf{x},y\in\mathcal{B}_{T}}\mathcal{L}(f_{\boldsymbol{\theta}}(\mathbf{x}),\mathbf{y})$
		\STATE Compute gradient $\mathbf{g}_{M}=\nabla_{\boldsymbol{\theta}}\sum_{\mathbf{x},y\in\mathcal{B}_{M}}\mathcal{L}(f_{\boldsymbol{\theta}}(\mathbf{x}),\mathbf{y})$
		\FOR{$I$ iterations}
		\STATE Update composite memory\\$\mathcal{M}\leftarrow opt_{C}(D(\mathbf{g}_{T},\mathbf{g}_{M}),\gamma_{C},I)$
		\ENDFOR
		\STATE Sample from buffer $\mathcal{B}_{B}\sim\mathcal{B}$
		\FOR{$J$ iterations}
		\STATE Update DNN parameters\\$f_{\boldsymbol{\theta}}\leftarrow opt_{T}(\sum_{\mathbf{x},y\in\mathcal{B}_{B}\bigcap\mathcal{B}_{T}}\mathcal{L}(f_{\boldsymbol{\theta}}(\mathbf{x}),\mathbf{y}), \gamma_{T}, J)$
		\ENDFOR
		\ENDFOR
		\ENDFOR
		\STATE Update buffer $\mathcal{B}\leftarrow\mathcal{B}\bigcap\mathcal{M}$
	\end{algorithmic}
	\caption{Dataset condensation}
	\label{alg:condensation}
\end{algorithm}

\begin{algorithm}
	\begin{algorithmic}[1]
		\REQUIRE DNN $f_{\boldsymbol{\theta}}$, buffer $\mathcal{B}$, sequence of tasks $\mathcal{T}_{1\leq i\leq N}$, training iterations $S$, training batch size $B_{T}$, training optimizer $opt_{T}$, training learning rate $\gamma_{T}$
		\ENSURE Trained DNN $f_{\boldsymbol{\theta}}$
		\FOR{Each task $\mathcal{T}_{1\leq i\leq N}$}
		\STATE Initialize DNN $f_{\boldsymbol{\theta}}$
		\STATE Condense dataset according to Algorithm \ref{alg:condensation}
		\FOR{$S$ iterations}
		\STATE Sample minibatch from buffer $\mathcal{B}_{T}\sim\mathcal{B}$
		\STATE Update DNN parameters\\$f_{\boldsymbol{\theta}}\leftarrow opt_{T}(\sum_{\mathbf{x},y\in\mathcal{B}_{B}}\mathcal{L}(f_{\boldsymbol{\theta}}(\mathbf{x}),\mathbf{y}), \gamma_{T}, 1)$
		\ENDFOR
		\ENDFOR
	\end{algorithmic}
	\caption{Training procedure}
	\label{alg:training}
\end{algorithm}

\subsection{Algorithm}
\label{ssec:Algorithm}

For our method we combine both, dataset condensation as described in section \ref{ssec:DatasetCondensation} and our novel composite memory outlined in \ref{ssec:CompositeMemory}, and adopt them to the incremental class learning problem as stated in section \ref{sec:IncrementalClassLearning}. Similarly to all rehearsal based methods, we utilize a small buffer $\mathcal{B}$ in order to retain what was previously learned and avoid catastrophic forgetting. But instead of storing input-output pairs of a task that are selected directly from the training data, we use dataset condensation to condense all available data into a small set of synthetic samples. Algorithm \ref{alg:condensation} describes our approach to dataset condensation using a composite memory.

Given a task $\mathcal{T}_{i}$, we start by randomly initializing the composite memory for this task using a uniform distribution over the interval $\left[0,1\right]$ for the components $\mathbf{c}_{i}$ and a standard normal distribution for the mixing coefficients $w_{ij}$. Similar to \cite{zhao2020dataset} we then randomly initialize the DNN that is used for condensation in an outer loop as we intend to learn synthetic examples that are useful for many different initializations. Next we sample minibatches of data from the task that we want to condense and data from our composite memory. It is crucial for our method and the original dataset condensation that this data is sampled from one class only as condensing examples from multiple classes at the same time is much more difficult and leads to worse results. While this restriction might be quite different from standard DNN training, it fits naturally into the incremental class learning problem where only a few classes are present in each task. This means we run Algorithm \ref{alg:condensation} independently for each class of a given task. The sampled minibatches are then used to compute the corresponding gradients using the same DNN parameters $\boldsymbol{\theta}$ and loss function $\mathcal{L}$. Next the condensation optimizer $opt_{C}$ is used to minimize the distance between gradients computed on the training and synthetic data according to the distance $D(\cdot,\cdot)$ from Eq. \ref{eq:grad_matching}. We minimize this distance for $I$ steps with a step size of $\gamma_{C}$ in order to learn the components and mixing coefficients of our composite memory. Differently from the original dataset condensation proposed by Zhao et. al. we then sample examples from the rehearsal buffer $\mathcal{B}$, combine them with the data sampled in step $5$ and use the training optimizer $opt_{T}$ to update the DNNs parameters with a learning rate of $\gamma_{T}$ for $J$ iterations. This is done to more closely resemble the DNNs training with a small rehearsal buffer during condensation and obtain synthetic samples that behave well in incremental class learning. Additionally it avoids training the DNN using only data from the current task which contains only a small number of classes and therefore can lead to overfitting on them. Besides the composite memory this is another important difference between our method that is specifically designed for continual learning and the original dataset condensation. Finally the rehearsal buffer $\mathcal{B}$ is updated with the learned components and mixing coefficients.

The training procedure of our method as described in Algorithm \ref{alg:training} differs from naive rehearsal. Instead of training on the data of a task directly, we first condense it and then train on the synthetic samples. We do this in order to speed up training as there are much fewer synthetic examples than original ones. During our experiments we also noticed this leads to a more stable training and overall better results.

\section{Experiments}

\begin{table}
	\centering
	\caption{Hyper parameters used throughout all of our experiments.}
	\begin{tabular}{lc}
		\toprule
		\textbf{Parameter} & \textbf{Value}\\
		\midrule
		Training learning rate $\gamma_{T}$ & $0.01$\\
		Training iterations $S$ & $500$/$1000$\\
		Condensation learning rate $\gamma_{C}$ & $0.1$\\
		Outer iterations $K$ & $100$\\
		Inner iterations $T$ & $10$\\
		Matching iterations $I$ & $10$\\
		Training iterations $J$ & $1$\\
		Training batch size $B_{T}$ & $128$\\
		Condensation batch size $B_{C}$ & $256$\\
		\bottomrule
	\end{tabular}
	\label{tab:hyperparameter}
\end{table}

\begin{figure*}
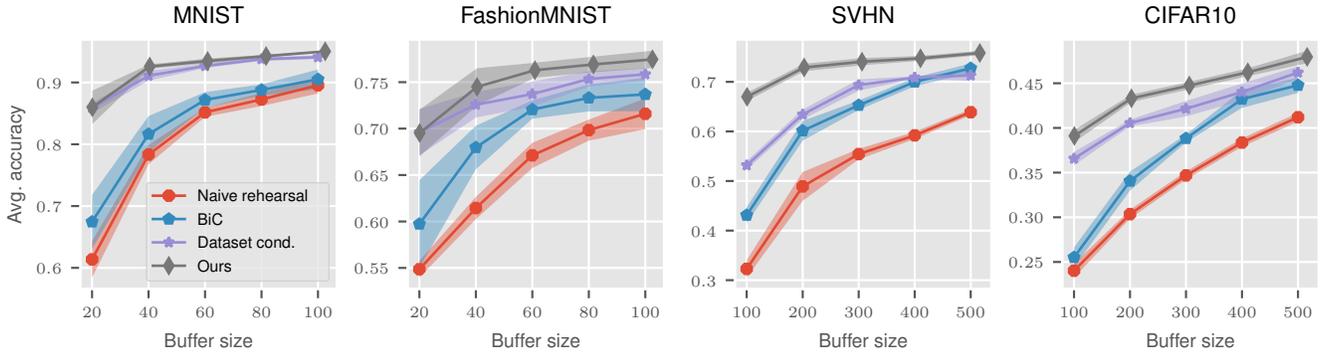

	\centering
	\include{./Figures/plot_results}
	\vspace{-0.75cm}
	\caption{Average accuracy and standard deviation of different methods on the tested datasets over five independent runs. For naive rehearsal and dataset condensation the buffer size is the total number of examples stored in the buffer after all tasks have been trained. For our method we convert the storage requirement into an equivalent number of examples in order to compare with the other methods.}
	\label{fig:avg_acc}
\end{figure*}

\subsection{Datasets}

In order to asses the effectiveness of our method we test it on a variety of datasets commonly used for continual learning. These are the MNIST and FashionMNIST dataset containing $28\times 28$ pixel gray scale images of digits and clothes as well as the SVHN and CIFAR10 datasets featuring $32\times 32$ pixel RGB images of house numbers from Google Street View and natural objects. All of these datasets have in common that they contain $10$ classes that can be split up into a sequence of tasks as described in section \ref{sec:IncrementalClassLearning}. For this we choose to split them into five tasks containing two classes in ascending order, i.e. $y_{j}^{1}\in\lbrace0,1\rbrace$, $y_{j}^{2}\in\lbrace2,3\rbrace$, $\ldots$, $y_{j}^{5}\in\lbrace8,9\rbrace$.

\subsection{DNN Architecture}

In our experiments we use the same DNN for all datasets. It is based on a modular architecture that is popular in the field of few-shot learning \cite{snell2017prototypical, vinyals2016matching} and was also used by Zhao et. al. for their experiments on dataset condensation. It consists of $3$ structurally identical blocks containing a convolutional layer with $128$ filters of size $3\times 3$ that is followed by an Instance Normalization layer \cite{DBLP:journals/corr/UlyanovVL16}, a ReLU activation and an average pooling block that averages blocks of size $2\times 2$. The output layer is a fully connected dense layer with Softmax activation. Similar to \cite{zhao2020dataset} our experiments suggest that normalization layers like InstanceNorm provide better results than none or batch normalization. We agree with their reasoning that the latter performs worse than other normalization techniques because batch statistics can not be estimated reliably for a small number of examples. Furthermore no data augmentation or other data pre-processing except normalization to the interval $\left[0,1\right]$ is applied to the inputs.

\subsection{Hyper parameters}
\label{ssec:Hyperparameters}

We choose the same hyper parameters for all experiments and avoided elaborate tuning of them. The only exception to this are the number of training and outer loop iterations $S$ and $K$. On the MNIST and FashionMNIST datasets a lower number of $500$ training and $50$ outer loop iterations are sufficient while on SVHN and CIFAR we double these in order to ensure convergence. Only a rough manual optimization was performed. During this we did not notice any particularly sensitive hyper parameters. The visual quality and performance increased with the number of iterations for all iteration parameters and an increase in the distillation batch size. This is to be expected as it leads to more and bigger mini batches being used during condensation. Our methods performance and its memory usage also depends strongly on the number of learned components $\mathbb{c}_{i}$ and weights $w_{ij}$. While we select the same number of components for our method as the other methods are allowed to store examples in their buffer, we learn twice the number of weights associated to them. This leads to a slightly higher storage requirement when we store the same number of components as the other methods store examples. TABLE \ref{tab:hyperparameter} contains a summary of the hyper parameters used in our experiments.

\subsection{Baselines}

We compare our method to several baselines and related methods. The simplest but still strong baseline is naive rehearsal of examples. For this, individual training samples are randomly selected and stored in the buffer. We select the same amount of examples per class in order to not introduce any bias towards a particular class. During training we randomly sample images from the rehearsal buffer and mix them with the same amount of training images from the new task in order to form a mini batch. This introduces a bias towards classes of the most recently trained task which is studied in more detail by \cite{wu2019large}. In order to correct for this bias the authors propose their BiC method which we include in our experiments as a recently proposed rehearsal based method for continual learning. As we are dealing with small buffer sizes we use all stored examples instead of keeping a separate validation set for BiC. Another baseline is the application of dataset condensation as a replacement of random selection in naive rehearsal. Here the original algorithm from \cite{zhao2020dataset} is changed slightly by including samples from the rehearsal buffer as described in section \ref{ssec:Algorithm}. To ensure a fair comparison of these methods with ours we use the same DNN architecture, shared hyper parameters and buffer size for all methods.

\subsection{Results}
\label{ssec:Results}

Fig. \ref{fig:avg_acc} shows the accuracy on a sequence of five tasks after the final task was trained. Numerical values are provided in TABLE \ref{tab:results}. Since we are interested in how the methods perform for different buffer sizes, we plot the average accuracy across all tasks over the number of examples stored in the buffer. A good performance for a small number of stored examples is generally desirable since it means more tasks can be stored in a fixed size buffer without compromising on performance. The average accuracy and its standard deviation is determined over five independent runs with different random initializations. Our method outperforms naive rehearsal and BiC by a large margin on all datasets. This is especially evident for small buffer sizes. On the less challenging dataset MNIST the gap between our method and the other methods closes quite quickly as the buffer size increases. This is due to the model reaching an average accuracy that is close to the upper bound of training on all tasks simultaneously. For the more challenging datasets, e.g. SVHN and CIFAR10, our method can maintain its advantage for much longer since on these datasets much more examples in the rehearsal buffer are needed to reach the upper bound. Our composite memory as described in section \ref{ssec:CompositeMemory} improves performance significantly when compared with the original dataset condensation. While there is a small but consistent improvement on MNIST for buffer sizes larger than $40$, a more significant increase in performance can be observed on the more challenging datasets FashionMNIST, SVHN and CIFAR10. We attribute this to dataset condensation being able to synthesize many more representative examples than simply storing them. This is due to the features of a particular class in these datasets being much more diverse than on MNIST whose classes are very narrowly defined. It is also interesting to note that our method using the composite memory is matched by dataset condensation for a buffer size of $20$ on MNIST and FashionMNIST. Here only one component per class is stored which seems to be insufficient for outperforming dataset condensation that can save two independent examples.

\subsection{Qualitative visual comparison}

\begin{figure*}
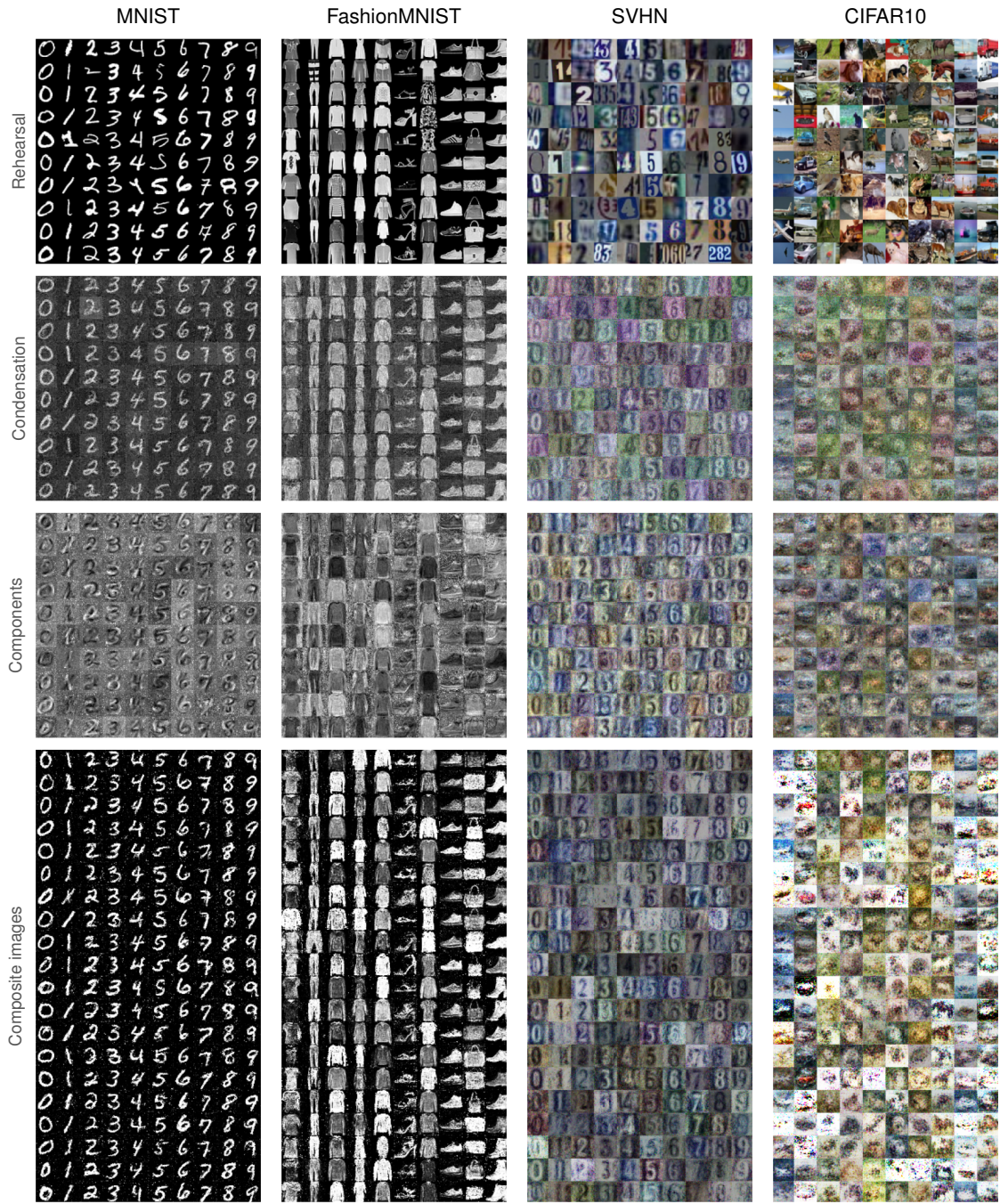

	\centering
	\include{./Figures/plot_imgs}
	\vspace{-0.75cm}
	\caption{Images stored in a buffer of size $100$ for naive rehearsal, dataset condensation and composite memory. The first two row shows examples used in naive rehearsal. Images learned using dataset condensation as proposed by \cite{zhao2020dataset} are shown in row two. The learned components of our method are shown in row three while the last row shows the $200$ composite images that are a weighted combination of the components above. Note that all images were normalized into the range $\left[0,1\right]$ for plotting which leads to a loss in contrast for dataset condensation images and components as they are learned and individual pixels of them can be outside this interval.}
	\label{fig:imgs}
\end{figure*}

Fig. \ref{fig:imgs} shows images stored in the buffer for naive rehearsal, dataset condensation and our method using the composite memory described in section \ref{ssec:CompositeMemory}. These are obtained at the end of training on the incremental class learning sequence when the buffer is completely filled and all classes are trained. When comparing images stored by naive rehearsal with dataset condensation and our method, there is a noticeable difference in visual clarity. Images that are learned, i.e. dataset condensation and our method, lack some detail and clearly defined textures. They also seem to be corrupted by noise. Although these images might be considered inadequate if the task we were trying to solve would be learning a generative model, they are arguably better images when it comes to using them for rehearsal in an incremental class learning setup. This is objectively quantified by the results discussed in section \ref{ssec:Results}. This can also be evaluated subjectively in case of the SVHN dataset. Here the images that are stored by naive rehearsal feature not only the centered digit that defines the class label but also neighboring digits of the house number. In the cases of methods that learn synthetic images for rehearsal, dataset condensation and our method, only the center digit is present. This indicates that the latter two methods learn only the important class defining features and ignore what is irrelevant for the task at hand. It is further interesting to compare dataset condensation with our method. The components learned using the composite memory seem to contain distinctive features of their class. But while some of them are bright others in the same image are dark. This can easily be seen in the case of FashionMNIST where some T-Shirts and Pullovers body has high and the arms have low intensity. This in addition to the weighted combination of all components results in much more diverse examples from our method when compared with dataset condensation. It is also interesting to note that images learned by dataset condensation lack diversity in their intensity.

\section{Conclusion}

We proposed a novel rehearsal based method for continual learning which is based on dataset condensation. For this a small set of synthetic training examples is learned by matching the gradients it produces during training with those from examples of the original dataset. In order to even further increase the performance and memory efficiency of rehearsal with dataset condensation, we propose a composite memory that shares features of a particular class among the learned synthetic examples. This leads to a significant increase in memory efficiency and performance when compared to other rehearsal based methods as we can share common features and therefore only need to store them once.

\begin{table}
	\caption{Average accuracy and standard deviation of our experiments. *Overhead applies only to our method and describes the additional storage requirement of the weights $w_{ij}$ measured in equivalent examples.}
	\label{tab:results}
	\centering
	\begin{tabular}{lccccc}
		\toprule
		Buffer size & $20$ & $40$ & $60$ & $80$ & $100$\\
		Overhead* & $0.10$ & $0.41$ & $0.92$ & $1.63$ & $2.55$\\
		\midrule
		& \multicolumn{5}{c}{\textbf{MNIST}}\\
		Naive rehearsal & $61.4$ & $78.3$ & $85.2$ & $87.3$ & $89.5$\\
		& $\pm2.9$ & $\pm1.46$ & $\pm0.7$ & $\pm1.1$ & $\pm1.3$\\
		BiC & $67.4$ & $81.6$ & $87.2$ & $88.8$ & $90.5$\\
		& $\pm4.4$ & $\pm2.9$ & $\pm1.3$ & $\pm0.9$ & $\pm1.6$\\
		Dataset cond. & $85.9$ & $91.1$ & $92.7$ & $93.8$ & $94.1$\\
		& $\pm0.8$ & $\pm0.8$ & $\pm0.3$ & $\pm0.3$ & $\pm0.3$\\
		Ours & $\mathbf{86.0}$ & $\mathbf{92.6}$ & $\mathbf{93.5}$ & $\mathbf{94.3}$ & $\mathbf{95.0}$\\
		& $\mathbf{\pm2.7}$ & $\mathbf{\pm0.4}$ & $\mathbf{\pm0.4}$ & $\mathbf{\pm0.2}$ & $\mathbf{\pm0.1}$\\
	\end{tabular}
	\begin{tabular}{lccccc}
		\\ & \multicolumn{5}{c}{\textbf{FashionMNIST}}\\
		Naive rehearsal & $54.9$ & $61.4$ & $67.1$ & $69.8$ & $71.6$\\
		& $\pm0.9$ & $\pm1.2$ & $\pm1.4$ & $\pm1.1$ & $\pm1.7$\\
		BiC & $59.7$ & $68.0$ & $72.1$ & $73.3$ & $73.7$\\
		& $\pm4.7$ & $\pm2.3$ & $\pm1.0$ & $\pm1.5$ & $\pm1.8$\\
		Dataset cond. & $\mathbf{69.6}$ & $72.6$ & $73.7$ & $75.8$ & $76.1$\\
		& $\mathbf{\pm2.5}$ & $\pm1.4$ & $\pm1.3$ & $\pm0.8$ & $\pm0.7$\\
		Ours & $69.5$ & $\mathbf{74.5}$ & $\mathbf{76.3}$ & $\mathbf{77.0}$ & $\mathbf{77.5}$\\
		& $\pm2.5$ & $\mathbf{\pm2.0}$ & $\mathbf{\pm0.8}$ & $\mathbf{\pm0.8}$ & $\mathbf{\pm0.9}$\\
		\bottomrule
	\end{tabular}
	\begin{tabular}{lccccc}
		\toprule
		Buffer size & $100$ & $200$ & $300$ & $400$ & $500$\\
		Overhead* & $0.65´$ & $2.60$ & $5.86$ & $10.4$ & $16.3$\\
		\midrule
		& \multicolumn{5}{c}{\textbf{SVHN}}\\
		Naive rehearsal & $32.3$ & $48.9$ & $55.4$ & $59.2$ & $63.9$\\
		& $\pm1.8$ & $\pm2.9$ & $\pm1.0$ & $\pm0.7$ & $\pm0.6$\\
		BiC & $43.1$ & $60.1$ & $65.3$ & $70.0$ & $72.7$\\
		& $\pm1.6$ & $\pm2.0$ & $\pm0.9$ & $\pm0.9$ & $\pm1.0$\\
		Dataset cond. & $53.2$ & $63.4$ & $69.4$ & $70.9$ & $71.3$\\
		& $\pm0.8$ & $\pm1.0$ & $\pm1.1$ & $\pm0.3$ & $\pm1.1$\\
		Ours & $\mathbf{67.0}$ & $\mathbf{72.9}$ & $\mathbf{74.1}$ & $\mathbf{74.7}$ & $\mathbf{75.8}$\\
		& $\mathbf{\pm0.8}$ & $\mathbf{\pm0.7}$ & $\mathbf{\pm0.7}$ & $\mathbf{\pm0.5}$ & $\mathbf{\pm0.5}$\\
	\end{tabular}
	\begin{tabular}{lccccc}
		\\ & \multicolumn{5}{c}{\textbf{CIFAR10}}\\
		Naive rehearsal & $24.0$ & $30.4$ & $34.7$ & $38.4$ & $41.2$\\
		& $\pm0.8$ & $\pm0.5$ & $\pm0.5$ & $\pm0.5$ & $\pm0.5$\\
		BiC & $25.5$ & $34.1$ & $38.8$ & $43.2$ & $44.8$\\
		& $\pm1.1$ & $\pm1.1$ & $\pm0.4$ & $\pm1.0$ & $\pm0.9$\\
		Dataset cond. & $36.6$ & $40.5$ & $42.1$ & $44.0$ & $46.2$\\
		& $\pm0.7$ & $\pm0.4$ & $\pm0.9$ & $\pm0.9$ & $\pm0.7$\\
		Ours & $\mathbf{39.1}$ & $\mathbf{43.3}$ & $\mathbf{44.7}$ & $\mathbf{46.2}$ & $\mathbf{47.9}$\\
		& $\mathbf{\pm0.7}$ & $\mathbf{\pm0.4}$ & $\mathbf{\pm0.4}$ & $\mathbf{\pm0.4}$ & $\mathbf{\pm0.7}$\\
		\bottomrule
	\end{tabular}
\end{table}

\bibliographystyle{IEEEtran}
\bibliography{./refs.bib}

\end{document}